\documentclass{article}
\usepackage[utf8]{inputenc}
\usepackage{authblk}
\usepackage{float}
\usepackage{xcolor}
\usepackage{tikz}
\usepackage{soul}
\usepackage{smartdiagram}
\usepackage{caption}
\usepackage{arev}
\usetikzlibrary{mindmap}
\usepackage{pdflscape}
\usepackage{hyperref}
\usepackage{subcaption}
\usepackage{graphicx}
\usepackage{comment}
\usepackage{multirow}
\usepackage{lineno}
\usepackage{array}
\usepackage{xcolor}
\usepackage{lscape}
\usepackage[sorting=none]{biblatex}
\usepackage{setspace}
\DeclareUnicodeCharacter{0301}{\'{o}}
\usepackage{amssymb}
\usepackage[a4paper, margin=1in]{geometry}
\usepackage{times}

\makeatletter
\renewcommand{\maketitle}{\bgroup\setlength{\parindent}{0pt}
\begin{flushleft}
  \textbf{\@title}

  \@author
\end{flushleft}\egroup
}
\makeatother

\providecommand{\keywords}[1]
{
  \textbf{\textit{Keywords---}} #1
}

\title{Are Linear Regression Models White Box and Interpretable?}
\date{}
\author[1,2,3,4]{Ahmed M Salih}
\author[1]{Yuhe Wang}
\affil[1]{Department of Population Health Sciences, University of Leicester, University Rd, LE1 7RH, Leicester, UK}
\affil[2]{William Harvey Research Institute, NIHR Barts Biomedical Research Centre, Queen Mary University of London, Charterhouse Square, London, EC1M 6BQ, London, UK}
\affil[3]{Barts Heart Centre, St Bartholomew’s Hospital, Barts Health NHS Trust, West Smithfield, London, EC1A 7BE, UK}
\affil[4]{Department of Computer Science, University of Zakho, Duhok road, Zakho, Kurdistan, Iraq}

\begin{document}
\maketitle
\thispagestyle{empty} 

\noindent

\begin{abstract}
\noindent
Explainable artificial intelligence (XAI) is a set of tools and algorithms that applied or embedded to machine learning models to understand and interpret the models. They are recommended especially for complex or advanced models including deep neural network because they are not interpretable from human point of view. On the other hand, simple models including linear regression are easy to implement, has less computational complexity and easy to visualize the output. The common notion in the literature that simple models including linear regression are considered as "white box" because they are more interpretable and easier to understand. This is based on the idea that linear regression models have several favorable outcomes including the effect of the features in the model and whether they affect positively or negatively toward model output. Moreover, uncertainty of the model can be measured or estimated using the confidence interval. However, we argue that this perception is not accurate and linear regression models are not easy to interpret neither easy to understand considering common XAI metrics and possible challenges might face. This includes linearity, local explanation, multicollinearity, covariates, normalization, uncertainty, features contribution and fairness. Consequently, we recommend the so-called simple models should be treated equally to complex models when it comes to explainability and interpretability. 
\end{abstract}
\newpage

\section{Introduction}
Explainable artificial intelligence (XAI) emerged to help understanding how machine learning models work. It uses set of tools and algorithms to convert the complex models into a more digestible form from human prospective~\cite{xu2019explainable}. XAI has several favorable aims and outputs including features attribution, fairness, uncertainty and sensitivity~\cite{salih2023review}.\\
Machine learning models are somehow grouped into two groups, simple and complex models. Complex models include deep neural network and conventional neural network while simple models include linear regression and rule-based model. It is recommended that XAI should be applied precisely to complex models because they are considered as~\textit{"black box"}~\cite{salih2023explainable}. On the other hand, simple models are easy to implement, fast in terms of running time compared to complex models and easy to visualize. Simple models including linear regression models (LRMs) are considered as self-explanatory and they are interpretable by their nature~\cite{dwivedi2023explainable}~\cite{arrieta2020explainable}~\cite{love2023explainable}. As opposed to \textit{"black box"} models, LRMs are called \textit{"white box"} models because they are understandable~\cite{loyola2019black}. This notion is common in the literature based on the fact that simple models including LRMs are easy to interpret because they provide several outputs which helps to understand the internal mechanism of the model. For instance, it shows the effect size of each explanatory variable in the model toward the model prediction by providing the coefficient value. Moreover, the sign of the coefficient value helps to determine whether the effect positive or negative~\cite{uyanik2013study}. In addition, it shows the uncertainty of the estimated coefficient value by reporting the confidence interval of each independent variable.\\
However, there are many challenges and obstacles that hinder to explain and interpret those models precisely when they are employed with real-life applications. This includes how to interpret the coefficient value when the variables are collinear, how to deal with the impacts of covariates and explain their effect with the current approaches to mitigate their impacts on model prediction. In addition, to what extent it is trustworthy to consider confidence interval as a proxy to measure the uncertainty of the model. Moreover, how to deal with the pre-processing steps including normalization and standardization which they make it more difficult to interpret the LRMs. The current paper discusses and sheds the light on some challenges related to explain LRMs. In the following sections and sub-sections, we will discuss some challenges  related to interpret and explain LRMs.
\section{Linear Regression Models}
LRMs are the most common and the simplest method to reveal the association between an explanatory variable and a continues outcome~\cite{hope2020linear}. The input data or variable is called the independent variable while the outcome/output is called the dependent variable. Equation~\ref{SLR} shows a simple linear regression when there is one dependent variable and an outcome.
\begin{equation}
    y \simeq  \beta_{0} + \beta X + \epsilon
\end{equation}\label{SLR}
where $y$ is the output, $\beta_{0}$ is the intercept, $\beta$ is the coefficient value (slope) of the explanatory variable $X$ and $\epsilon$ is the error which represents the difference between the predicted value and the actual value. The $\beta$ value represents the effect of one unit of $X$ toward $y$. 
LRMs could involve more than one independent variable which is called multiple linear regression. Equation~\ref{MLR} shows an example of a multiple linear regression model when the number of independent variables up to $n$ where $n$ could be any number.
\begin{equation}
    y \simeq \beta_{0} + \beta_{1}X_{1}+\beta_{2}X_{2}+ \beta_{3}X_{3} + ...... + \beta_{n}X_{n} + \epsilon
\end{equation}\label{MLR}
In this case there are multiple $\beta$ which each represents the effect/coefficient value of its own independent variable. The interpretation of the $\beta$ represents the effect in one unit of the variable of interest toward the outcome while assuming/holding all other independent variables in the model constant~\cite{nimon2013understanding}. LRMs are extended to predict categorical outcome such as sex, death and a disease~\cite{castro2022linear} with classification models including logistic regression. In this case, the outcome of the model is a probability (between 0 and 1) from a sigmoid function indicating whether it belongs to a specific class or not~\cite{castro2022linear}.\\
From statistical point of view, LRMs are used to perform association between an independent variable or group of independent variables and an outcome. On the other hand, LRMs from machine learning point of view are used to predict an outcome using sets of independent variables. Comparing to advanced and complex models, LRMs are considered as \textit{white box} and interpretable based on the perception that the internal mechanism of these models is understandable. 
\section{XAI}
XAI is a set of tools, approaches, methods or algorithms that help ends users to understand how the model works, how it is reached to a specific decision, what are the most informative variables in the model and to what degree the model is certain~\cite{salih2024perspective}. Such aims are very important to build trust and increase transparency to better implement AI models in real life applications. Moreover, it helps to improve model performance and mitigate bias effect in the model. There have been many XAI proposed in the recent years to explain and interpret AI models globally for all instances in the model or locally for a specific instance. In addition, XAI could be either model-agnostic which means can be applied to any model or model-specific indicating that can be applied to a specific model. Moreover, some XAI methods are proposed for a specific kind of data. For instance, Grad-CAM~\cite{selvaraju2017grad} and Integrated gradients~\cite{sundararajan2017axiomatic} are proposed to explain AI models with image data; Accumulative Local Effect~\cite{apley2020visualizing} and Partial Dependency Plot~\cite{greenwell2018simple} are applied with tabular data while Shapley Additive Explanations~\cite{lundberg2017unified} and Local Interpretable Model-agnostic Explanations~\cite{ribeiro2016should} can be applied to both imaging and tabular data.\\
XAI has been proposed and recommended to be applied to complex models including deep learning, conventional neural network and  recurrent neural network. This is because these models are considered as~\textit{"black boxy"} because the internal mechanism of these model is not clear. On the other hand, simple models including LRMs are considered more interpretable and easier to understand. It is indeed the common notion in the literature that these models are self-explanatory with less recommendation to employ XAI models.
\section{Challenges of Interpreting LRMs}
The following subsections discuss the most common challenges that end-users might face when interpret LRMs apart from which domain.
\subsection{Linearity assumption}
\begin{figure}[H]
\centering
\includegraphics[height=5 cm, keepaspectratio]{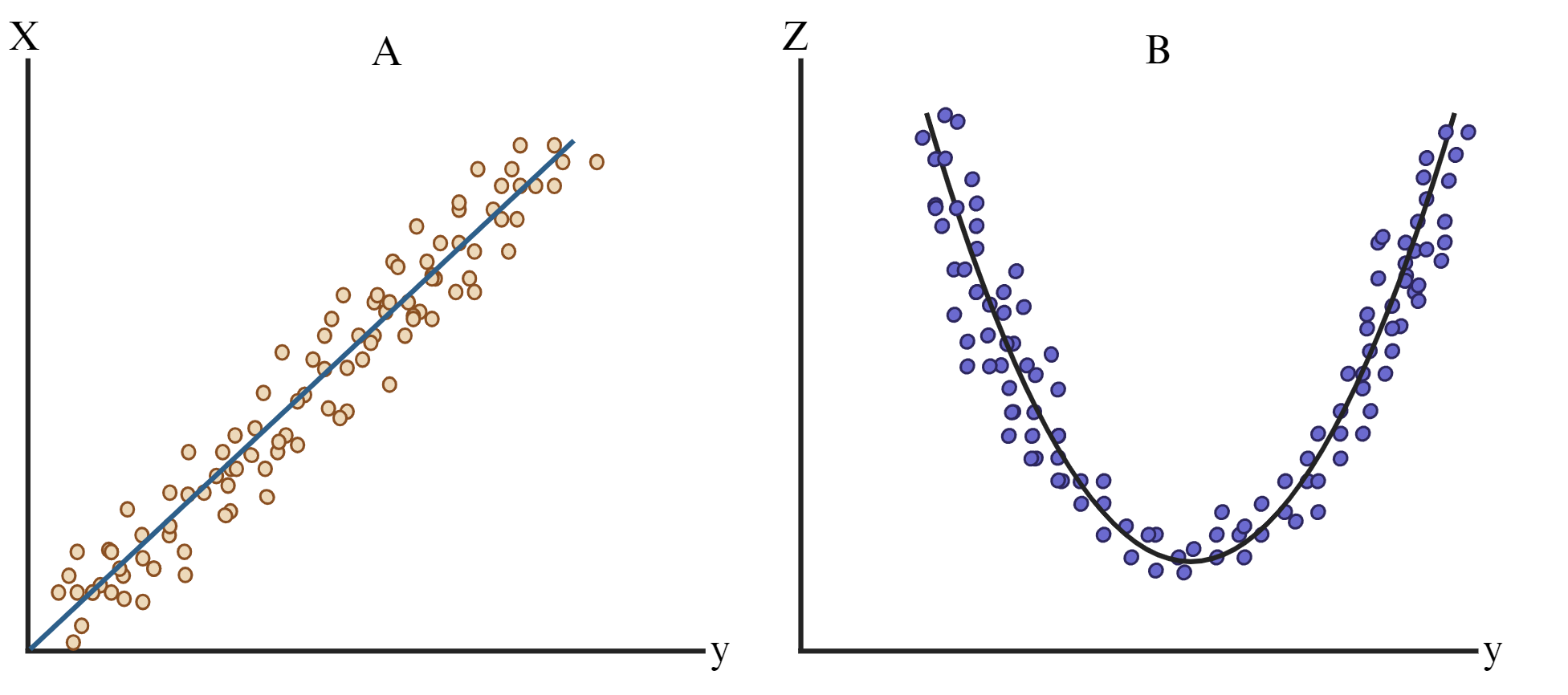}
\caption{Linear vs non-linear association.}\label{Linearity}
\end{figure}

As their names imply, LRMs assume and enforce a linear association between the independent variables and the output. The assumption is based on the theory background behind these models. Figure~\ref{Linearity}shows the association could be linear or non-linear. The association between X and y is a positive linear association while it is non-linear between Z and y. It is important to understand the association between the input and the output in order to use an appropriate model and interpret it accordingly.
\noindent
In real-world applications the association between the input variables and output might be linear, monotonic or more complex. One of the significant aims of XAI is to reveal the kind of the association between the input data and the output. For example, the association between the number of reservations of booking rooms in a European touristy city and the temperature degree. Such association might not be linear and better represented by a U-shape. This is because the association between the number of reservations and temperature degree would increase to a point (mid of summer) and then turn around and decreases. To explain and interpret the LRMs with this kind of data, the end-users report the coefficient value as the effect size and the direction of the effect which is not accurate and does not reflect the actual association.
\subsection{Local explanation}
\begin{figure}[H]
\centering
\includegraphics[height=5 cm, keepaspectratio]{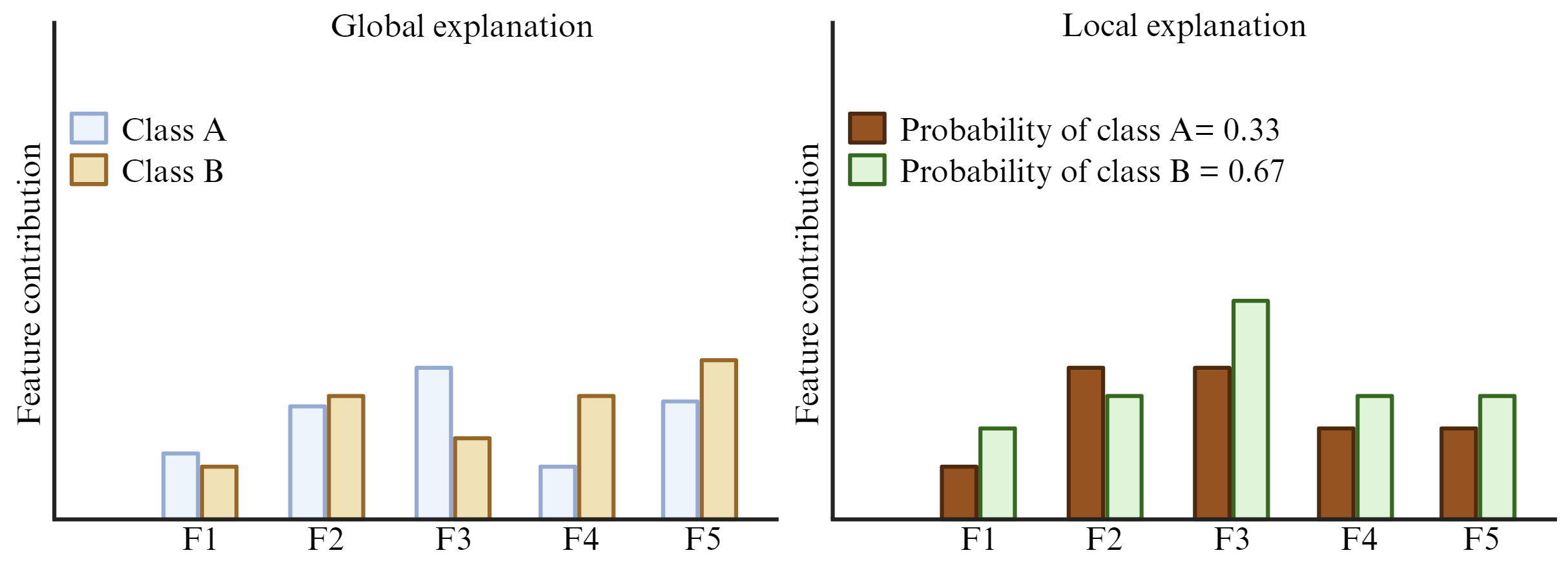}
\caption{Global explanation vs local explanation.}\label{LocalXAI}
\end{figure}

Machine learning models can be explained and interpreted either globally or locally. Globally indicates that to explain the model at global level considering all samples in the model. On the other hands, local explanation means explain the model locally for a specific instance in the model~\cite{salih2023explainable}. In other word, it means showing the effect of the explanatory variables for one sample. Figure~\ref{LocalXAI} shows a global and a local explanation. It shows the contribution of each feature toward any class at individual level and the probability for being class A or B.
\noindent
It is very important to show the explanation at individual level because the effect of each independent variable is averaged over all samples in the global explanation. The local explanation of an instance might be different from the explanation at global level. This kind of explanation is more vital in real life applications. For instance, a client would like to know what information in the mortgage application form made his/her application weak and eventually the mortgage was rejected. Unfortunately, LRMs are lack of such valuable property which is one of the most significant aims of XAI methods. LRMs within the context of machine learning only provides the coefficient value from the training datasets. This value is an average effect considering all samples in the model. As a result, LRMs cannot be considered as a white box within this context because it is not clear how it works at local level.
\subsection{Multicollinearity}
\begin{figure}[H]
\centering
\includegraphics[height=5 cm, keepaspectratio]{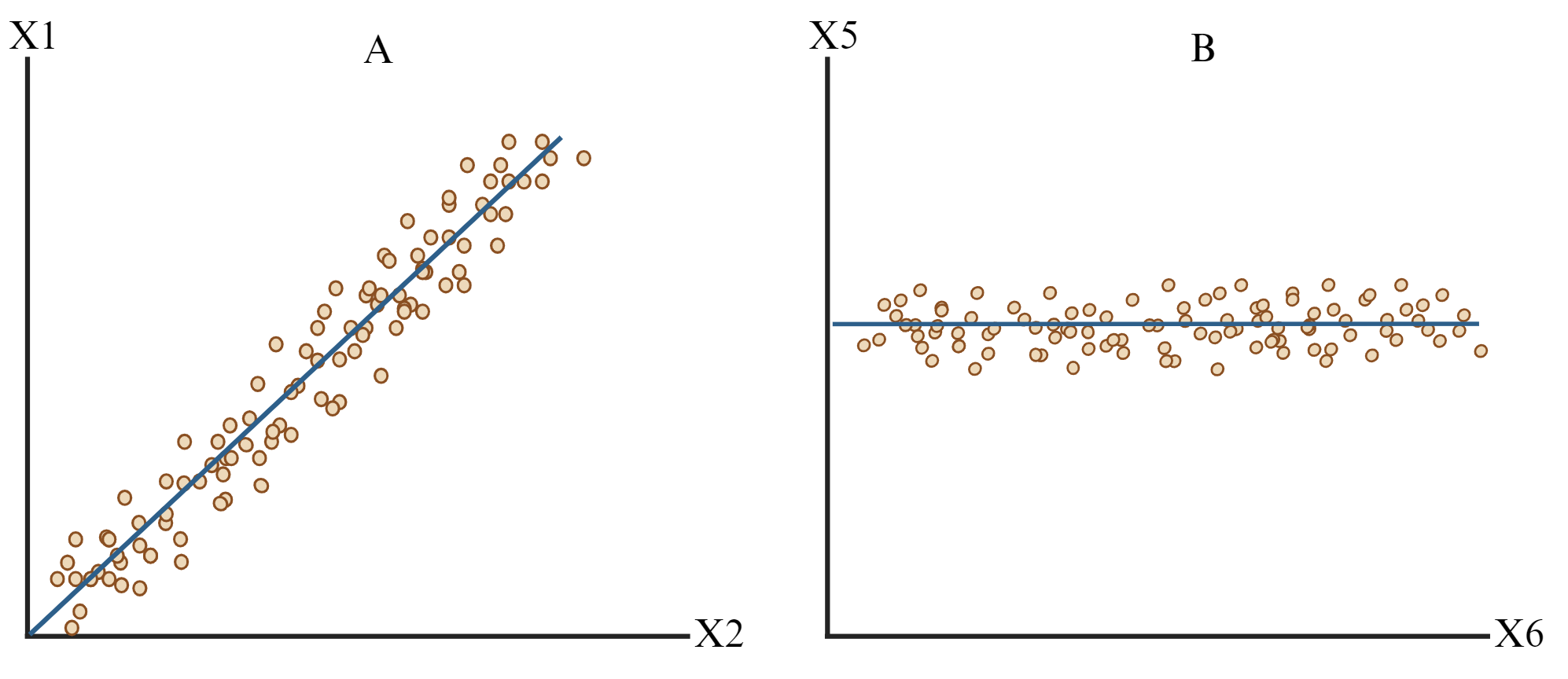}
\caption{Multicollinearity and independent features.}\label{Collinear}
\end{figure}

Multicollinearity is one of the common phenomena in statistical analysis when two or multiple independent variables in the model are highly correlated~\cite{alin2010multicollinearity}. In other word, it indicates when some of explanatory variables are linear function of the others in the model. Figure~\ref{Collinear} shows that X1 and X2 are highly positively correlated (A). On the other hand, it shows (B) that there is no correlation or dependency between X5 and X6 which indicates the absence of collinearity. It is more evident in real-life applications especially in health care, biology and medicine. To interpret the LRMs, the coefficient value of each variable in the model is reported alongside the confidence interval as the impact on the dependent variable. It shows the effect size and the direction of the effect whether it is positive or negative toward the model prediction. Coefficient value is considered one of the main properties of LRMs that makes it interpretable.
\noindent
The interpretation of the coefficient value represents the effect of one unit of the independent variable of interest on the outcome while holding all other independent variables in the model constant~\cite{ziglari2017interpreting}. Such interpretation might be correct when the independent variables are really independent (as the case between X5 and X6 in figure~\ref{Collinear}). However, in real life applications the independent variables are usually collinear and they change simultaneously. Accordingly, the classic interpretation of the coefficient value is not realistic and cannot be considered to explain and interpret the LRMs model. One might argue that features selection could be applied to select none-correlated features before feeding them to the model. Thereafter, the classic interpretation of the coefficient value accurately explains the model. In most cases especially in health care domain, researchers would like to include all the features in the model because each one has different clinical interpretation and might result in different recommendation. 
\subsection{Covariates}
\begin{figure}[H]
\centering
\includegraphics[height=5 cm, keepaspectratio]{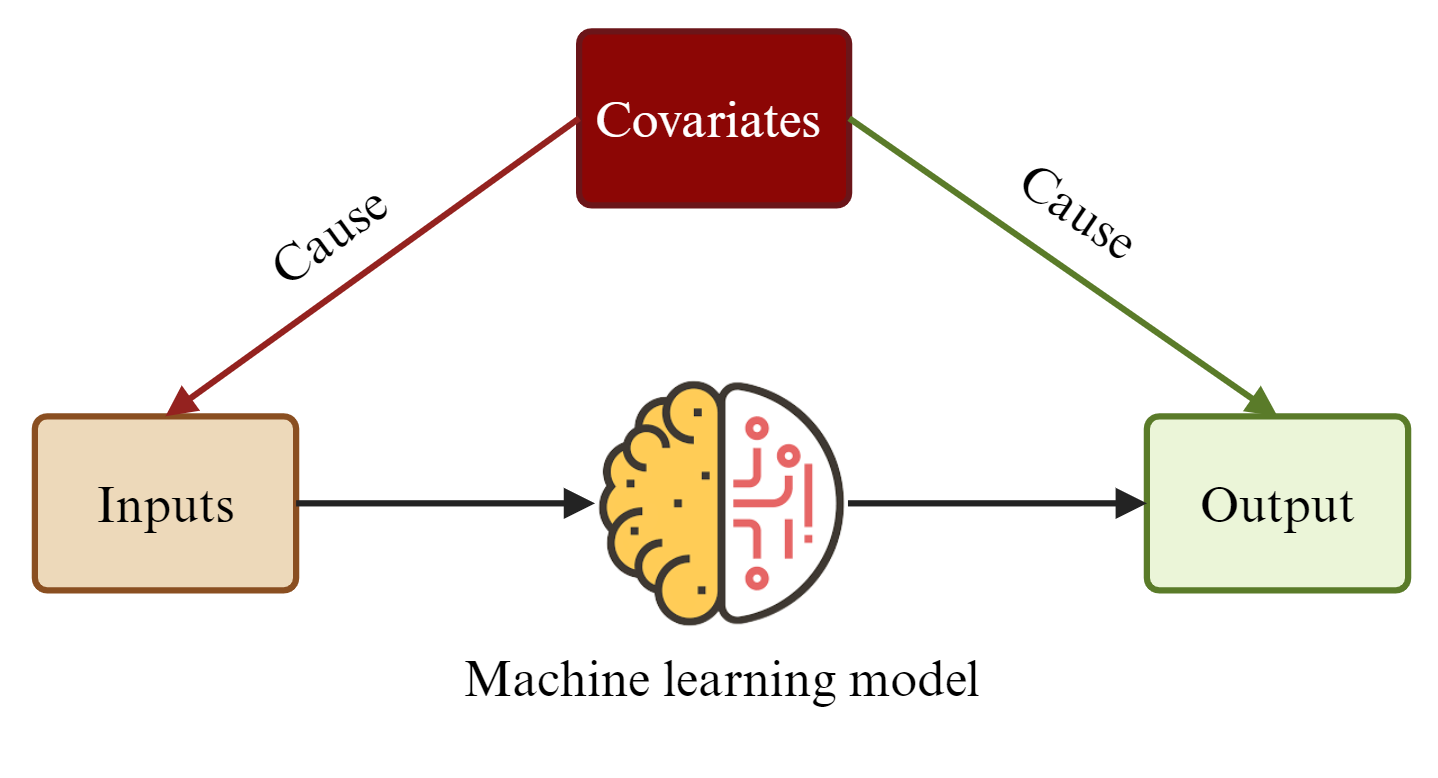}
\caption{The association of covariates with the input and the output.}\label{Covariats}
\end{figure}

The covariates are group of variables or factors that affect both the independent and dependent variables in the model simultaneously~\cite{zhao2020training}. As it is shown in figure~\ref{Covariats}, the Covariates have direct causal association to the model input and output at the same time. The covariates are different from a domain to another and might be related to characteristics of the samples or data acquisition. For example, sex, ethnicity and age are common covariates in healthcare applications. In addition, weight and height or body mass index are common covariates in cardiovascular diseases. To mitigate the effect of covariates, different approaches are proposed including regressing the covariates from the independent variables before feeding them to the model. In addition, some researches use the covariates directly in the model as independent variable. Another approach is to intentionally select cases and control using some matching methods such as propensity score based on the set of observed covariates~\cite{zhao2020training}.
\noindent
When it comes to explain and interpret the LRMs, it is challenging on how to reveal the impacts of those factors. Let us consider a scenario where we want to predict a disease using electronic health records as independent variables. In such case, sex, ethnicity and age might be considered as observed covariates. When regressing the covariates from the independent variables, we embed the impact of the covariates in the independent variables. In such case, it is not possible to to explain how the model would behave when it is applied to only male or female or on other ethnicity. For instance, the effect of sex is regressed from independent variables. However, in some domains including cardiovascular and brain diseases male are more prone to experience those diseases than female~\cite{bots2017sex}~\cite{wingo2023sex}. In this case, it is very vital to explain and interpret the model when switching between sexes. Similar issue will appear if we consider propensity score to match samples because we just naturalize the independent variables over the set of covariates. The last scenario is when including the covariates in the model alongside the independent variable. The classical interpretation of the coefficient value in the model makes it more difficult to interpret the model because it enforces holding the other independent variables in the model as constant. Consequently, it is indeed difficult to interpret the LRMs and consider the impact of covariates in the three discussed approaches.
\subsection{Data scaling}
\begin{figure}[H]
\centering
\includegraphics[height=5 cm, keepaspectratio]{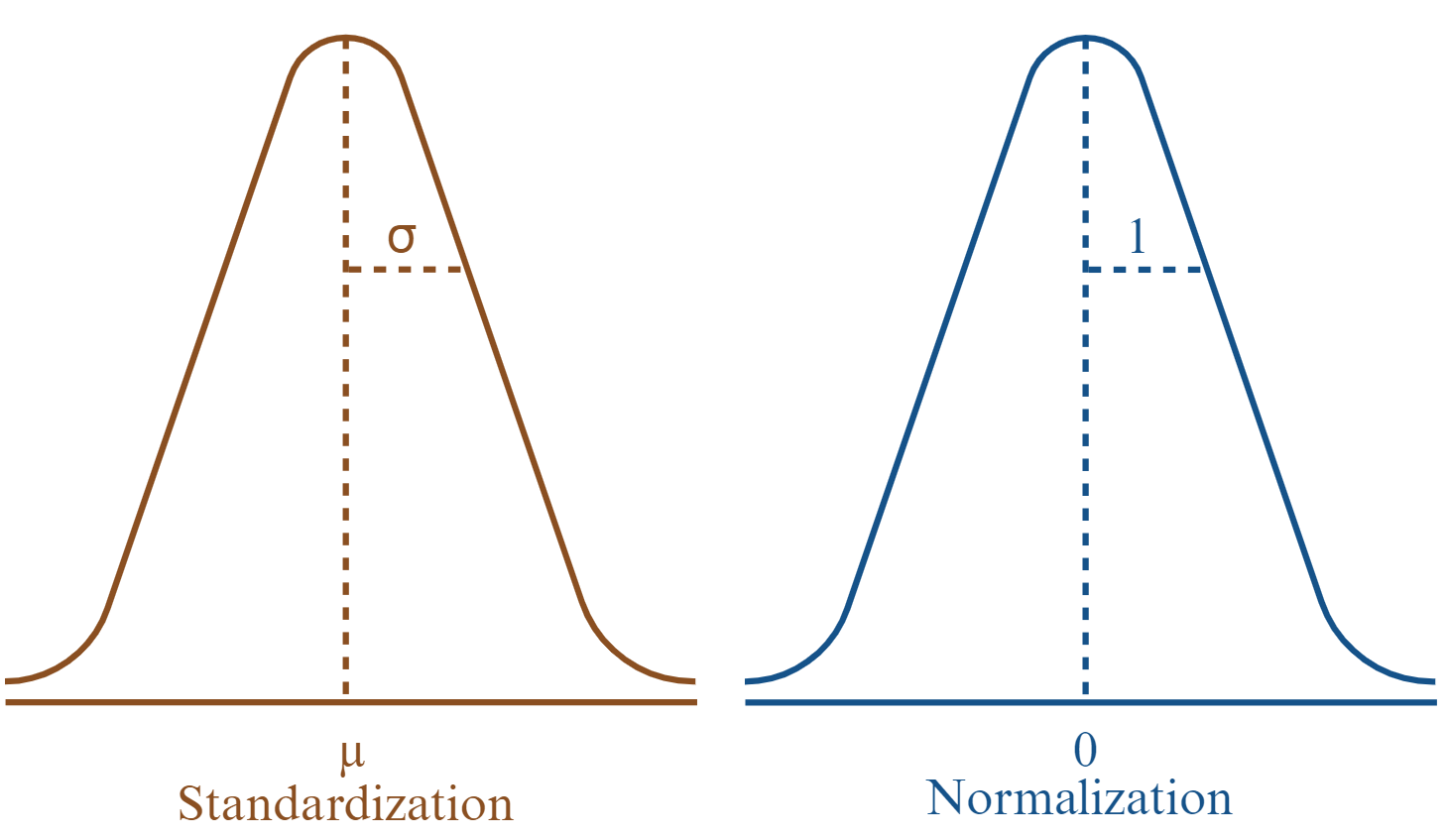}
\caption{Data pre-processing.}\label{Scaling}
\end{figure}

Usually, the data are not in the same scale or they might not have similar distribution. Some of them might have a wide range while others are very tight. Those with wider range and higher values might dominant model decision. One of the most common step of data pre-processing in machine learning is to either normalize or standardize the data before fitting them in the model~\cite{vieira2020main}. As it is explained in figure~\ref{Scaling}, normalization is the process of scaling the data to have same range which is usually between zero and one while standardization means convert the data to have zero mean and one unit standard deviation. It helps to improve model performance, decrease the running time of the model and allowing the model deals with all data equally.\\
\noindent
However, such process hinders the ability to explain and interpret the model by revealing the effect size in unit. For instance, the normalization method converts the features into unitless. Consequently, the common interpretation of the coefficient value is no longer possible because it does represent change in unit. Similarly, the coefficient value of the standardized data represents the deviation from the center of the data which is usually the mean value. In both cases, the end users will not be able to understand how change in the independent variable would lead to effect on the outcome. The only way to interpret the model is to compare the coefficient values of all independent variables to recognize the one with higher and smaller effect. However, such interpretation still not precise because it does not consider the multicollinearity phenomena.
\subsection{Uncertainty}
\begin{figure}[H]
\centering
\includegraphics[height=3 cm, keepaspectratio]{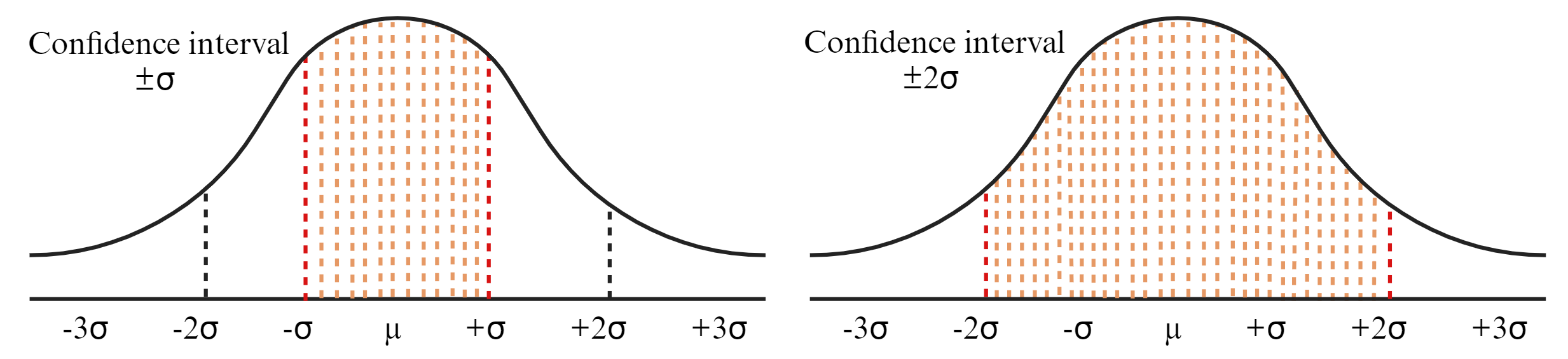}
\caption{confidence interval of two estimates.}\label{CI}
\end{figure}

One of the desirable properties of XAI aims is to show the uncertainty in the model when making a prediction~\cite{alufaisan2021does}. Uncertainty helps the end users whether to consider a specific prediction or not. Moreover, it might be one of the core stone to trust a model and employ it daily life applications. Uncertainty could be presented and shown in many shapes. For instance, in classification models the probability of a subject belong to a specific class is a form of uncertainty. Moreover, confidence interval of the estimated coefficient value is another form of uncertainty. Figure~\ref{CI} shows the confidence interval of two estimated coefficient value. The left part of the figure shows the confidence interval is one unit ($\pm$) of standard deviation away from the mean while the one on the right shows that the confidence interval is two units ($\pm$) standard deviation away from the mean. The width of the confidence interval is considered as a proxy of precision of the estimated value. The more the confidence interval small, the more the model certain of the estimate. On contrary, the larger the confidence interval, the less the model certain.
\noindent
Although confidence interval can be calculated in LRMs, the main question is how to specify a threshold to identify good certainty. Moreover, the confidence interval has many drawbacks as it is explained in~\cite{morey2016fallacy}. Richard et al. show that there are many fallacies related to how interpret the confident interval including that it can be considered as a measure of certainty~\cite{morey2016fallacy}. Moreover, they show in practical examples that the width of the confidence interval does not represent an index of precision of an estimate nor the plausibility of the estimate value. Accordingly, using the confidence interval as a measure of certainty to interpret and explain LRMs is not accurate and might be biased.
\subsection{Features contribution in classes}
\begin{figure}[H]
\centering
\includegraphics[height=5 cm, keepaspectratio]{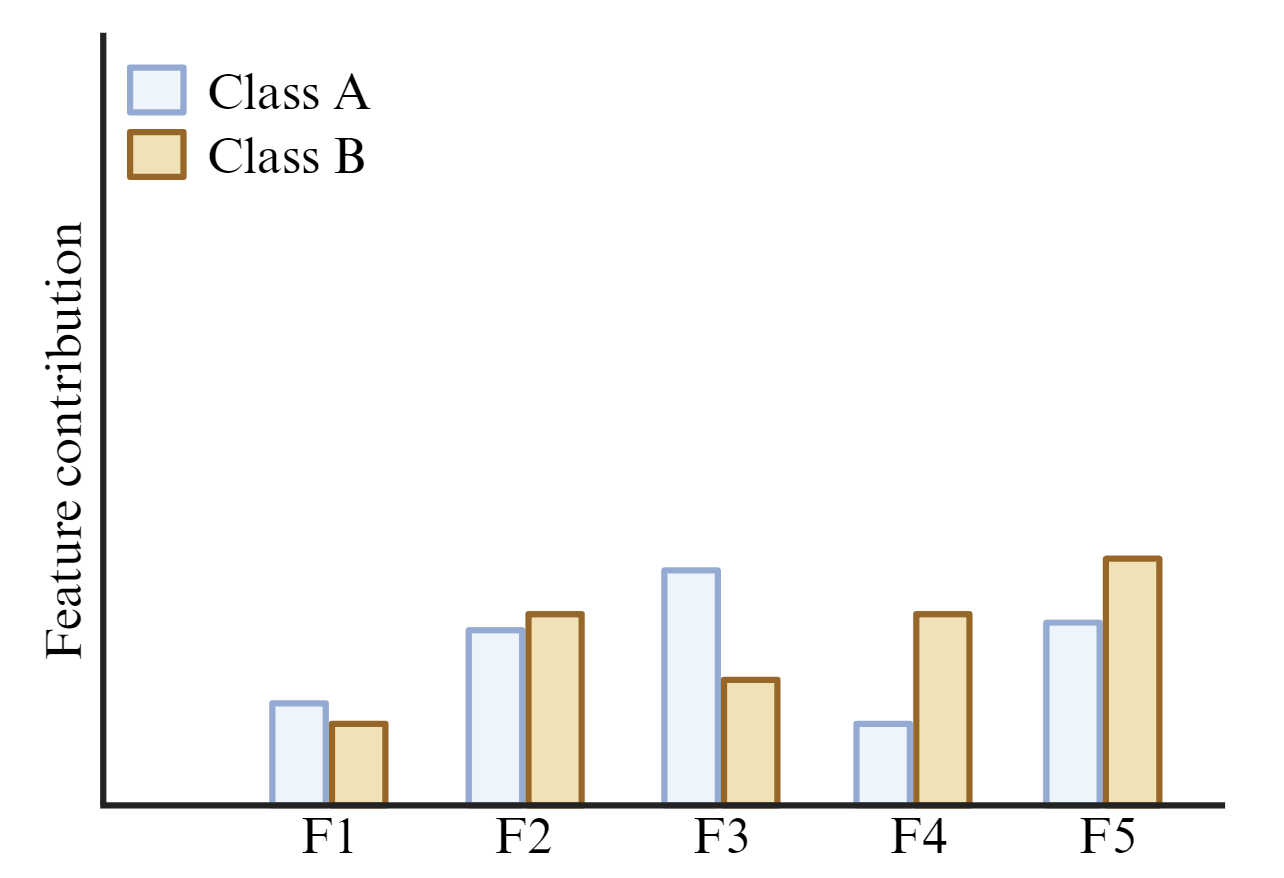}
\caption{Features contribution in each class}\label{FAC}
\end{figure}

Regression models are used to perform binary classification as the case with logistic regression. The most common scenario is to model the outcome (cases vs control) using a set of independent variables~\cite{sperandei2014understanding}. It is vital to interpret the model in both cases when the outcome is either one of the classes. For instance, figure~\ref{FAC} as an example of interpreting a model to reveal the contribution of the features in each class. It shows that some features contribute more in class A compared to their contributions in class B. This is very important to understand the model and might help in features engineering or even before to collect the data. 
\noindent
Features contribution is one of the most common output of XAI method especially with tabular data~\cite{dwivedi2023explainable}. It shows the contribution of the features toward the prediction and their impacts in each class when modelling a classification task. Unfortunately, LRMs do not provide information or explain how the independent variables contribute in each class. For instance, they do not show which features contribute more significantly in class A or B. Moreover, such information cannot be shaped from LRMs neither at global level for all subjects nor at local level for a specific instance. Consequently, LRMs are lack of features attribution for each class which is one of the significant components of XAI outcome.
\subsection{Fairness}
\begin{figure}[H]
\centering
\includegraphics[height=6 cm, keepaspectratio]{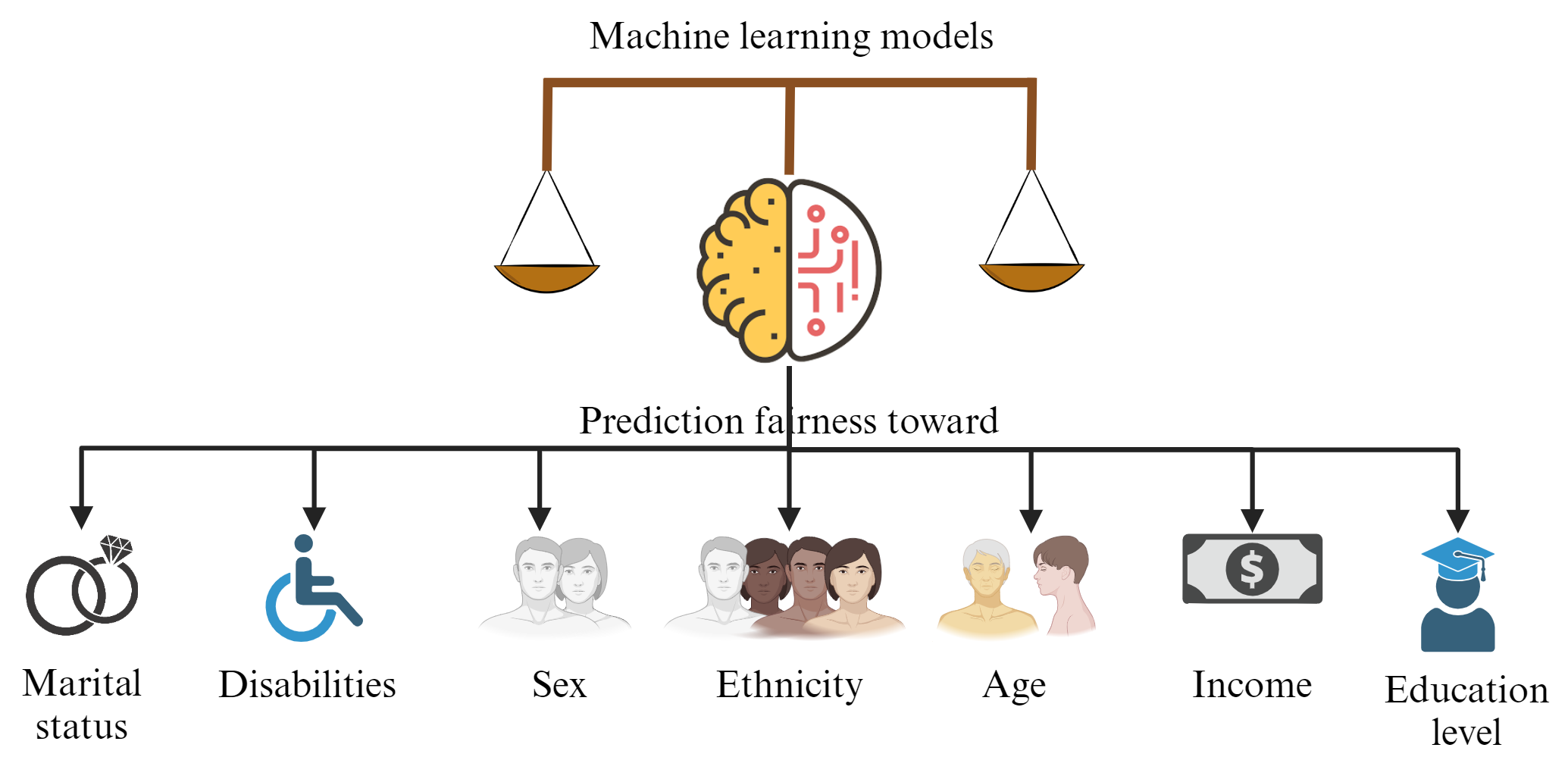}
\caption{Fairness towards sub-groups.}\label{Fairness}
\end{figure}

There are many proxies proposed to measure the trustworthy of machine learning models including LRMs. Fairness one of the most common proxy that machine learning models should fulfil especially in health care and medicine. Machine learning algorithms including LRMs are usually applied to data that involve sub-groups. For instance, the data might include both sexes, multiple races or ethnicity, disabilities, education level, income, marital status or data collected from young and old participants. Figure~\ref{Fairness} shows that machine learning models should be fair toward these groups when making a prediction. In other word, the prediction should not be biased toward a specific group or class~\cite{mehrabi2021survey}. There has been some metrics proposed to measure the fairness of a model. For instance, the equal opportunity metric indicates that a model is fair if the true positive rate (TPR) is equal for subgroups in a binary classifier model. Equal opportunity is extended to equal odd by ensure that both TPR and false positive rate (FPR) should be equal in subgroups.~\cite{hardt2016equality}.
\noindent
LRMs by themselves are not fair and biased against minorities in the data. This phenomenon is intrinsically inherited in LRMs even if the data from the minorities included in the training step~\cite{cohen2023fairness}. Furthermore, most of the proposed metrics to measure fairness of any model are suitable for classification models when true positive and false positive can be calculated. Accordingly, it is more challenging to measure fairness in LRMs. More importantly, it might be impossible to satisfy all the aspects of fairness and maximize the accuracy of the model at the same time~\cite{berk2021fairness}. 
\section{Open issue and recommendations}
Although LRMs are used widely, easy to implement, has less computational complexity and does not require massive data pre-processing steps, they hold some challenges which make them difficult to explain and interpret. The common notion in the literature that LRMs are more interpretable and less accurate compared to complex models including deep neural network is not precise. In the aforementioned points, we showed that LRMs are quite hard to interpret, explain and they might have the same challenges to interpret advanced models. LRMs should be treated as complex models when it comes to explainability and interpretability. It is true that all the points we mentioned could be applied to all machine learning models and not limited to LRMs. However, we focused on LRMs to defy the common notion of interpretability related to LRMs. We believe the following recommendations should be considered to ensure a more accurate interpretation of LRMs.
\begin{enumerate}
    \item LRMs should be treated equally to complex models when it comes to interpretability and explainability. This applies to explain the model at global and local level.

    \item Coefficient value can be user to explain and interpret the effect size and direction of the input data when there is no collinearity. However, it is not a precise proxy to explain the model as effect size when LRMs are applied to collinear data or the association is not linear.
    
    \item Multicollinearity is one of the main issues to explain any machine learning model no matter whether it is simple or complex. End users might dig deeper in the literature to find possible solutions and suggestions to mitigate its effect~\cite{salih2024explainable}.

    \item Covariates need to be considered carefully when interpreting their impacts on the prediction outcome by revealing their causal association~\cite{holzinger2019causability}.

    \item Confidence interval might be not accurate to measure the uncertainty of a model. More sophisticated metrics including estimating distribution rather than a single point might be a better way to quantify the certainty~\cite{raftery1997bayesian}.

    \item Post-hoc XAI methods are indispensable to reveal the contribution of the features in each class when it is applied to a classification task because LRMs are lack of such property.

    \item To the best of our knowledge the issue of normalization and standardization regarding how to interpret the effect size has not been investigated before, nor proper approaches are proposed. More researches are required in this direction to ensure the possibility to apply the pre-processing steps without losing the ability to interpret the effect size in its original unit.

    \item LRMs are not fair and might be biased against or toward a specific group in the model. Several approaches and metrics were proposed to measure and improve fairness, specifically for LRMs~\cite{agarwal2019fair}~\cite{calders2013controlling}~\cite{berk2021fairness}. 
    
\end{enumerate}

\section{Acknowledgments}
AMS acknowledges support from The Leicester City Football Club (LCFC)

\printbibliography
\end{document}